\documentclass[letterpaper]{article} 
\usepackage{aaai25}  
\usepackage{times}  
\usepackage{helvet}  
\usepackage{courier}  
\usepackage[hyphens]{url}  
\usepackage{graphicx} 
\usepackage{natbib}  
\usepackage{caption} 
\usepackage{dblfloatfix}
\usepackage[utf8]{inputenc} 
\usepackage[T1]{fontenc}    
\usepackage{url}            
\usepackage{booktabs}       
\usepackage{amsfonts}       
\usepackage{nicefrac}       
\usepackage{microtype}      
\usepackage{lipsum}
\usepackage{amsmath}
\usepackage{fancyhdr}       
\usepackage{graphicx}       
\usepackage{tabularx}
\usepackage{array}
\usepackage{placeins}
\usepackage{float}
\usepackage{multicol}
\usepackage{cellspace} 
\addparagraphcolumntypes{X}
\usepackage{enumitem}

\title{A Unifying Information-theoretic Perspective on Evaluating Generative Models}
\author {
    Alexis Fox\textsuperscript{\rm 1},
    Samarth Swarup\textsuperscript{\rm 2},
    Abhijin Adiga\textsuperscript{\rm 2}
}
\affiliations {
    \textsuperscript{\rm 1}Duke University\\
    \textsuperscript{\rm 2}University of Virginia\\
    alexis.fox@duke.edu, swarup@virginia.edu, abhijin@virginia.edu
}

\usepackage{bibentry}

\newcommand{\knn}{\textit{k}NN }

\newcommand{\kinn}{\textit{k}NN}

\newcommand{\pce}{\mathit{PCE}}
\newcommand{\rce}{\mathit{RCE}}
\newcommand{\re}{\mathit{RE}}
\newcommand{\rr}{\mathcal{R}}
\newcommand{\rg}{\mathcal{G}}
\newcommand{\numX}{N_{X}}
\newcommand{\numY}{N_{Y}}
\newcommand{\numR}{N_{R}}
\newcommand{\numG}{N_{G}}
\newcommand{\ball}[4]{\#\big(#4, B_{#3, #1}(#2_i)\big)}
\newcommand{\ballone}[4]{\#\big(#4, B_{#3, #1}(#2_1)\big)}
\newcommand{\ballunit}[3]{B_{#1, #2}(#3_i)}
\newcommand{\frechet}{Fréchet }

\newcommand{\fd}{$FD$}

\newcommand{\thickhline}{\noalign{\hrule height 1pt}}

\begin{document}

\maketitle

\begin{abstract}
Considering the difficulty of interpreting generative model output, there is significant current research focused on determining meaningful evaluation metrics. Several recent approaches utilize ``precision'' and ``recall,'' borrowed from the classification domain, to individually quantify the output \textit{fidelity} (realism) and output \textit{diversity} (representation of the real data variation), respectively. With the increase in metric proposals, there is a need for a unifying perspective, allowing for easier comparison and 
clearer explanation of their benefits and drawbacks.
To this end, we unify a class of \textit{k}th-nearest neighbors (\textit{k}NN)-based metrics under an information-theoretic lens using approaches from 
\textit{k}NN density estimation. Additionally, we propose a tri-dimensional metric composed of Precision Cross-Entropy (PCE), Recall Cross-Entropy (RCE), and Recall Entropy (RE), which separately measure \textit{fidelity} and two distinct aspects of diversity, \textit{inter-} and \textit{intra-class}. Our domain-agnostic metric, derived from the information-theoretic concepts of entropy and cross-entropy, can be dissected for both sample- and mode-level analysis. Our detailed experimental results demonstrate the sensitivity of our metric components to their respective qualities and reveal undesirable behaviors of other metrics. 
\end{abstract}

\begin{links}
     \link{Code}{https://github.com/NSSAC/PrecisionRecallMetric}
     \link{Extended version}{https://arxiv.org/abs/2412.14340}
\end{links}

%

\section{Introduction}

\textbf{Need for Metrics.} Deep generative models, including Generative Adversarial Networks (GANs) \cite{goodfellow2014generativeadversarialnetworks}, Variational Autoencoders (VAEs) \cite{kingma2022autoencodingvariationalbayes}, and diffusion models \cite{sohldickstein2015deepunsupervisedlearningusing}, are achieving unprecedented realism in their outputs~\cite{ravuri2023understandingdeepgenerativemodels}. 
Their success highlights an urgent need for robust methods to evaluate their output quality. 
Unlike discriminative models, which are evaluated against a clear ground truth in the form of labeled test sets, generative models produce a spectrum of plausible outputs based on the learned data distribution. Hence, evaluating generative models is a nontrivial task. 

\textbf{Traditional 1D Metrics.} Often, an important goal of generative models is to produce outputs that human observers find realistic \cite{stein2024exposing}. While human assessment can be quantified, manual human evaluation is expensive, time-consuming, and makes continuous assessments impractical. 
Automatic evaluation metrics such as Inception Score ($IS$) \cite{salimans2016improved} and \frechet Inception Distance ($FID$) \cite{heusel2017gans} address these inefficiencies by computing the distributional differences of generated and real content. 

\begin{figure}[t]
\includegraphics{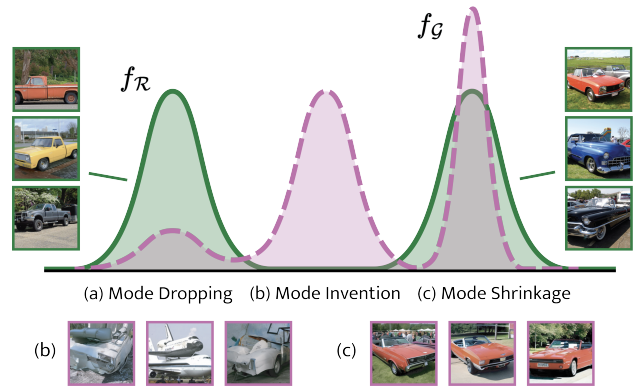}
\centering
\caption{We visualize the various failure modes (a)-(c) for a model distribution, where the real (green) distribution is composed of two modes: sport cars and pickup trucks. (a) represents a lack of generated pickup trucks, (b) and its corresponding examples show unrealistic points, and (c) depicts points clustered at the mode average. }
\end{figure}

\textbf{Motivation for 2D Metrics.} However, as one-dimensional scores, $IS$ and $FID$ cannot distinguish various shortcomings, e.g. when outputs are generated in low-probability regions of the real distribution (low fidelity) or insufficiently represent high-probability regions (low diversity) \cite{sajjadi18precision}. A recent set of two-dimensional metrics, which can be broadly categorized as versions of ``precision'' and ``recall,'' has emerged to differentiate the several types of failure modes. \citet{sajjadi18precision} introduce the first of a series of precision and recall metrics based on decomposing the real and generated distributions into common and unique components. Subsequent works propose alternatives to precision and recall, with most utilizing \textit{k}-Nearest Neighbor (\textit{k}NN) methods~\cite{kynkaanniemi19precision, naeem20fidelity, alaa22synthetic, cheema2023precision, Park_2023}.

\textbf{The Gap.} Despite the proliferation of evaluation methods, there remains a need for clearer theoretical comparisons and validation of these metrics. We address this gap by providing probabilistic derivations with information-theoretic interpretations for a subset of metrics, focusing on Precision Recall Cover \cite{cheema2023precision} and Density \& Coverage \cite{naeem20fidelity}.  

We also introduce a simple new information-theoretic metric that, unlike existing metrics, simultaneously distinguishes between the three failure modes visualized in Figure 1. Mode invention (1b) indicates a loss of precision (\textit{fidelity}), occurring when the model generates implausible outputs that do not align with the real distribution. Mode dropping (1a) represents one type of recall (\emph{diversity}) loss, where the model lacks sufficient coverage of certain real regions. Another recall issue, mode shrinkage (1c), happens when a model repeatedly produces highly similar points---lacking \textit{intra-class diversity}---that represent a mode ``average.''
We summarize our three main contributions below.

\textbf{Unifying Existing Metrics.} In Section~\ref{sec:unifying}, we provide information-theoretic derivations for Precision Recall Cover and Density \& Coverage. By extending lemmas from \citet{noshad2017direct}, we derive a bias term for the expected value of these metrics, allowing for easier interpretation and comparison and motivating our metric. 

\textbf{Novel Information-Theoretic Metric.} We propose a new metric for generative models, composed of Precision Cross-Entropy ($\pce$), Recall Cross-Entropy ($\rce$), and Recall Entropy ($\re$), based on the estimator from~\citet{leonenko2008class}. Unlike previous work, we separate the two types of diversity loss. Section~\ref{sec:our_metrics} details the definition of these scores as well as how they discern the various failure modes mentioned. We also describe how our population-level metric can be dissected for both mode- and sample-level analysis, useful for diagnostic purposes. 

\textbf{Experimental Results.} We list desiderata for generative model metrics in Section~\ref{sec:metrics_char}. Through our experimental results in Section~\ref{sec:experiments}, we show how our metrics align with these ideals, as well as how other existing metrics are less effective with respect to one or more of these. Specifically, we analyze the sensitivity of the measures to the different failure modes mentioned above, as well as examine their correlation with human scoring. As with most prior work, we focus on image evaluation, but our methods can be easily generalized to other data modalities, as seen in the extended version's Appendix.

\section{Background and Related Work}
Several metrics have been proposed to quantify the performance of generative models. Below we summarize relevant metrics and their respective benefits and shortcomings. We start with common background information. 

\textbf{Background}. Calculating distances in raw input spaces, such as image pixels, is often impractical due to the high dimensionality and sensitivity to noise. To address this, many metrics --- including ours --- embed data into a feature space that provides a more semantic representation, where Euclidean distances correlate better with human-perceived similarity. 
While expensive, human perception can be used as a ``ground truth'' for other metrics for realism. One benchmark, Human eYe Perceptual Evaluation ($\mathit{HYPE}$), measures human error rate when identifying fake and real images, serving as a proxy for realism.~\cite{zhou2019hype}.

\textbf{Classic Metrics}. Inception Score and log-likelihood have historically been popular metrics but are criticized for poor correlation with human judgment.~\cite{theis2015note, kolchinski2019approximating}. Fréchet Inception Distance ($FID$), which we later compare experimentally, is another widely used metric for evaluating generated images. $FID$ evaluates images by comparing the \frechet distance of generated and real images in an Inception-based feature space, modeled as multivariate Gaussians~\cite{heusel2017gans}. We use  the generalized abbreviation $FD$ to refer to the \frechet distance metric hereafter, as justified in Section 6. $FD$ is consistent with human judgment~\cite{stein2024exposing}; however, it assumes Gaussian-distributed features, which may not always hold~\cite{borji2022pros}.
Additionally, as a one-dimensional metric, $FD$ cannot distinguish failure modes, which is a motivation for the following multi-dimensional metrics.

\textbf{Early Notion of Precision and Recall}.
\citet{sajjadi18precision} introduce the novel Precision-Recall for Distributions ($\mathit{PRD}$) metric, which compares embedded samples from the real and generated distribution, using clustering assignments to assess whether the differences in their common support should be attributed to precision (accuracy) or recall (diversity) loss. Recognizing the limitations of the $\mathit{PRD}$ metric, which relies on relative densities and yields a continuum of values along a PR-curve, \citet{kynkaanniemi19precision} propose an Improved version of Precision and Recall ($IP$ \& $IR$) leveraging \knn distances of samples embedded in a feature space. $IP$ measures precision by assessing if each generated image is within the real image manifold, estimated with the union of \textit{k}NN balls  centered on each image. Symmetrically, $IR$ evaluates if each real image is within the generated image manifold.

\textbf{Recent Extensions}. Several subsequent works can be seen as extensions of the approaches by \citet{sajjadi18precision} 
and \citet{kynkaanniemi19precision}. While superficially different, most of these build upon common underlying
computational ideas. We focus in particular on the work of~\citet{cheema2023precision}, who propose Precision Coverage and 
Recall Coverage ($PC$ \& $RC$, collectively defined as $\mathit{PRC}$), which differs from previous works by parameterizing what is 
considered ``sufficient'' coverage, and on the work of \citet{naeem20fidelity}, who observed that $IP$ \& $IR$ are 
sensitive to outliers and fail to identify identical distributions. They developed the more robust Density and 
Coverage measures ($D$~\&~$C$). We focus on these two because they are either recent or have particular limitations (on which,
more in Section~\ref{sec:unifying}). However, other recent approaches can be seen in a similar information-theoretic 
light~\cite{alaa22synthetic, djolonga2020precision, liu21divergence}, which we comment upon briefly in the discussion. Outside the precision and recall framework, the Rarity Score \cite{han2022rarityscorenew} quantifies the uniqueness of individual generated samples, which, by comparing histograms, can be implicitly seen as a intra-class diversity measure.

We note that no metric discussed above simultaneously distinguishes between inter- and intra-class diversity, which is a strength of our metric detailed in Section~\ref{sec:our_metrics}.

\newcommand{\rd}{\mathcal{X}}
\newcommand{\gd}{\mathcal{Y}}

\section{Preliminaries}
Given a (multivariate) random variable~$\rd$ with distribution~$f_\rd$, let $X = \{X_1, \ldots, X_{N_X}\}$ be a set of $N_X$ i.i.d. samples 
of~$\rd$, with each \(X_i \in \mathbb{R}^d\).
In this work, we will frequently compare two distributions, letting~$f_\rd$ and~$f_\gd$ represent two general distributions, while~$f_\rr$ and~$f_\rg$ specifically represent the real and generated distributions, respectively.
Given sets of samples~$X$ and~$Y$ of $\rd$ and $\gd$, $Z$ 
is the combined set of points from $X$ and $Y$, denoted by $Z=X \cup Y$. 
\subsection{Information Theoretic Concepts}
\label{sec:concepts}
For completeness, we present some foundational concepts from information theory, which we will later integrate into the analysis of generative metrics.
\textbf{Shannon entropy} of a random variable~$\rd$ is defined 
as $H(\rd) = \mathbb{E}_{x\sim\rd} \log 
\frac{1}{f_\rd(x)}$ 
while \textbf{cross-entropy} of~$\rd$ relative to~$\gd$ is $CE(\rd, \gd) = \mathbb{E}_{x\sim\rd} \log \frac{1}{f_{\gd}(x)}$. \textbf{KL-divergence} of~$\rd$ from~$\gd$ is defined as $D_{KL}(\rd\|\gd) = \mathbb{E}_{x\sim\rd} \log  \frac{f_{\rd}(x)}{f_{\gd}(x)}$.

\textbf{R\'enyi divergence}, parameterized by order $\alpha$ ($\alpha \neq 1$ and $\alpha > 0$), is a generalization of KL-divergence that similarly quantifies the informational difference between two distributions \cite{renyi1961measures}. The formal definition is as follows:
\begin{align}\label{eqn:renyi_div}
D_{\alpha}(\rd \| \gd) & =\frac{1}{\alpha-1} \log \int f_{\rd}(x)^{\alpha} f_{\gd}(x)^{1-\alpha} d x \\
& =\frac{1}{\alpha-1} \log J_{\alpha}(f_{\rd}, f_{\gd}), 
\end{align}
where $J_{\alpha}(\rd, \gd)=\mathbb{E}_{x\sim\gd}\Big[\big(\frac{f_{\rd}(x)}
{f_{\gd}(x)}\big)^{\alpha}\Big]$ \cite{noshad2017direct}.  As $\alpha\rightarrow1$, 
R\'enyi divergence approaches the KL-divergence. We note that $\alpha = 1$ is defined for $J_{\alpha}(\rd, \gd)$. 

\subsection{\kinn-based Estimators}
\label{sec:knn}
In practical applications, the full distribution of $\rd$ or $\gd$ is often not known, or the integrals required to compute measures such as entropy or divergence are intractable. As a result, direct computation of these measures is not feasible, necessitating the estimation of these values from available data. Multiple methods~\cite{noshad2017direct, poczos2011estimation, leonenko2008class} have been proposed  that utilize the \textit{k}th-nearest neighbors of sample points to estimate probability density ratios. Below we introduce notation and definitions for such estimators. 

\textbf{\textit{k}NN Density Estimation.} 
Let $\ballunit{k}{X}{X}$ be a hypersphere in $\mathbb{R}^d$ centered at
the point $X_i$, with its radius determined by the distance to the $k$th-nearest neighbor in the set $X$.
Both the radius and number of points within a \knn ball can be used for density estimation. We define $\ball{X}{X}{k}{Y}$ as the number of points from set $Y$ found within the ball $\ballunit{k}{X}{X}$. To accommodate the distance-based nature of other estimators, we define $D_{k, Y}(X_i)$ as the distance from $X_i$ to its \textit{k}th-nearest neighbor in the
set $Y$.

\textbf{Entropy and Cross-Entropy Estimator.} 
\citet{leonenko2008class} 
propose direct, nonparametric estimators for
cross-entropy and entropy. These estimators leverage \knn distances as a
proxy for probability density. 

The estimators for entropy, $\breve{H}_k(X)$, and cross-entropy, $\breve{CE}_k(X, Y)$,  are defined below:
{\small
\begin{align}
\breve{H}_k(X) = \frac{1}{N_X} \sum_{i=1}^{N_X} \log \Big((N_X-1) 
e^{-\Psi(k)} \bar{c} \big( D_{k, X}(X_i)\big)^{d} \Big) \\
\breve{CE}_k(X, Y) = \frac{1}{N_X} \sum_{i=1}^{N_X} \log \Big( N_Y 
e^{-\Psi(k)} \bar{c} \big( D_{k, Y}(X_i) \big)^{d} \Big).
\end{align}
}
Here, $\Psi(z)=\Gamma^{\prime}(z) / \Gamma(z)$ represents the digamma function and $\bar{c}$ is the volume of a unit ball in $\mathbb{R}^d$. 

\textbf{Divergence Estimator.} \citet{noshad2017direct} define an
estimator for Rényi divergence utilizing nearest neighbor ratios. 
Their method calculates the ratio of the number of points from $X$ to that of $Y$ found within the hypersphere, $\ballunit{k}{Z}{Y}$, to 
estimate $\frac{f_{\rd}(x)}{f_{\gd}(x)}$. \citet{noshad2017direct} derive the 
following estimators for $J_{\alpha}(\rd, \gd)$ and $D_{\alpha}(\rd \| \gd)$, with $\eta=\numY/ \numX$:
\begin{equation}
\widehat{J}_{\alpha}(X, Y)=\frac{\eta^{\alpha}}{\numY} 
\sum_{i=1}^{\numY}\Big(\frac{\ball{Z}{Y}{k}{X}}{\ball{Z}{Y}{k}{Y}+1}\Big)^{\alpha} \label{eqn:Jalpha}\\
\end{equation}
\begin{equation}
\widetilde{D}_{\alpha}(X, Y)=\frac{1}{(\alpha-1)} \log 
\widehat{J}_{\alpha}(X, Y).
\end{equation}

\section{Unifying Information Theoretic Perspective}
\label{sec:unifying}
Below we examine the formulae for two metrics---Precision Recall Cover and Density \& 
Coverage---in order to express them within a unified information theoretic perspective. We then 
use our results to motivate our metric. 

\subsection{Precision Recall Cover}
\label{sec:prc}
\citet{cheema2023precision} propose the Precision Recall Cover ($\mathit{PRC}$) metric, consisting of Precision Coverage ($PC$) and Recall Coverage ($RC$). For their population-level analogue, they define two parameters, $\alpha$ and $\beta$, which allow for flexibility in determining what is considered ``sufficiently'' covered for both generated and real regions. \citet{cheema2023precision} additionally provide a  \kinn-based empirical definition, which we describe below. 
 
\textbf{Empirical $(k, k')$-PRC Definition.} For two positive 
integers~$k=Ck'$ where~$C$ is a positive integer as well, let $\alpha = 
{k'}/{\numG}$ and $\beta = {k}/{\numR}$. The $(k, k')$-Precision 
Coverage of sample set $R$ with respect to $G$ is defined by constructing 
$k$- and $k'$-nearest neighbor balls over the sample sets: 
\begin{align} 
PC_{k, k'}(R, G) = \frac{1}{\numG} \sum_{i=1}^{\numG} \textbf{1}[\ball{G}{G}{k}{R}\ge k'],
\label{eqn:PC}
\end{align}
where~$\textbf{1}[\cdot]$ is the indicator variable of the specified event.
This method calculates the proportion of $k$NN balls in $G$ that contain at least $k'$ points from $R$, normalized by the total points in $G$. $(k, k')$-Recall Coverage is defined similarly:
\begin{align} 
RC_{k, k'}(R, G) = \frac{1}{\numR} \sum_{i=1}^{\numR} \textbf{1}[\ball{R}{R}{k}{G}\ge k'].
\label{eqn:RC}
\end{align}

\textbf{Connecting PRC}. To set up its definition to be compatible with our perspective, we rearrange the expression of $PC$. Results for $RC$ generalize symmetrically.
By definition of~$\#(\cdot)$, the quantity~$\ball{G}{G}{k}{G}$ refers to
the number of points from~$G$ within the $k$-nearest neighborhood of~$G_i$, which is exactly~$k$.
The term in Equation \ref{eqn:PC} can be expressed as:
\begin{align}
\ball{G}{G}{k}{R}\ge k' \Rightarrow \frac{\ball{G}{G}{k}{R}}{\ball{G}{G}{k}{G}}\ge \frac{1}{C}\,.
\end{align}
Therefore, noting that $\eta=\frac{\numG}{\numR}$,
{\small
\begin{align} \label{eqn:pceta}
PC_{k,k'}({R}, {G}) = \frac{1}{\numG} \sum_{i=1}^{\numG} \mathbf{1}\bigg[\frac{\eta\cdot\ball{G}{G}{k}{R}}{\ball{G}{G}{k}{G}}\ge \frac{\eta}{C}\bigg]
\end{align} 
}
By~(\ref{eqn:Jalpha}), with $\alpha=1$ and $U=R \cup G$: 
\begin{align}
\widehat{J}_{1}(R, G)=\frac{1}{\numG} \sum_{i=1}^{\numG}\frac{\eta\cdot\ball{U}{G}{k}{R}}{\ball{U}{G}{k}{G}+1}.
\label{eqn:J1hat}
\end{align}
We define an indicator function version of the above and its corresponding empirical version as:
{\small
\begin{align}
\widehat{J'_{1}}(R, G)=\frac{1}{\numG} \sum_{i=1}^{\numG}\textbf{1}\bigg[\frac{\eta\cdot\ball{U}{G}{k}{R}}{\ball{U}{G}{k}{G}+1} \ge \frac{\eta}{C}\bigg]
\label{eqn:J1prime}
\end{align}
}
\begin{equation}
J'_{1}(\rr, \rg)=\mathbb{E}_{x\sim\rg}\left[\textbf{1}\left[\frac{f_\rr(x)}{f_{\rg}(x)}\ge \frac{\eta}{C}\right]\right].
\label{eqn:J1primetheoretical}
\end{equation}
Since the indicator function is not Lipschitz continuous, as required by Lemma 3.2 of \citet{noshad2017direct}, we define a close sigmoid approximation, where $\sigma(x; \theta) = \frac{1}{1 + e^{-\theta (x - \frac{\eta}{C})}}$, with $\theta$ controlling the steepness of the sigmoid (higher $\theta$ approaching closer to the indicator function). 
As seen above, the empirical definition of $PC$ (Eqn.~\ref{eqn:pceta}) and the estimator $\widehat{J'_{1}}$ (Eqn.~\ref{eqn:J1prime}) differ in only two aspects: $PC$  calculates the nearest neighbors within the set $G$ instead of $U$, and $PC$ has no added 1 in the denominator. In the Appendix, we show how the described differences affect the bias. We also define an information theoretic intuition for $PC$ through rearrangement of~Eqn. \ref{eqn:J1primetheoretical} and defining $C_2 = \log(\frac{C}{\eta})$:  
\begin{align}
J'_{1}(\rr, \rg)=\mathbb{E}_{x\sim\rg}\bigg[\:\textbf{1}\Big[\log\left(\frac{f_{\mathcal{G}}(x)}{f_{\mathcal{R}}(x)}\right) \le C_2\Big]\:\bigg].
\end{align}
The interior expression checks if the number of extra bits needed to encode the generated data point using the optimal real encoding scheme instead of generated is no greater than $C_2$.
By symmetry, we can relate Recall Coverage to $J'_{1}(\rg, \rr).$ 

\subsection{Density \& Coverage}

\citet{naeem20fidelity} define their metric, called Density (not to be confused with probability density) \& Coverage, with the motivation of building upon the work of \citet{kynkaanniemi19precision}, by creating a version of precision \& recall more resistant to outliers. Their definitions are described below.  

\textbf{Density \& Coverage Definition}. Density, used to measure the fidelity of generated samples, can be defined as:
\begin{equation}
\text { Density }=\frac{1}{\numG} \sum_{i=1}^{\numR} 
\frac{\ball{R}{R}{k}{G}}{k}.
\end{equation}
Density counts the number of neighborhood spheres of real samples that contain the generated data point, normalized by $\numG$. Their recall counterpart, Coverage, can be defined as: 
\begin{equation}
\text { Coverage }=\frac{1}{\numR} \sum_{i=1}^{\numR} \textbf{1}\big[\#(G, B_{k,R}(R_i))\ge 1\big].
\end{equation}\citet{naeem20fidelity} base their recall measure on the real sample set, justified by its lower outlier presence compared to generated sets." Coverage averages the number of real \knn balls that contain one or more generated samples. 

\textbf{Connecting Density}. We rearrange the expression of Density to 
define it within our framework. Noting that $k=\ball{R}{R}{k}{R}$, 
\begin{align}
\text { Density }= \frac{1}{\numG} \sum_{i=1}^{\numR} \frac{\ball{R}{R}{k}{G}}{\ball{R}{R}{k}{R}}.
\end{align}
We then define $\widehat{J}_{1}(G, R)$ and its theoretical value ${J}_{1}(\rg, \rr)$:
\begin{equation}
\widehat{J}_{1}(G, R) = \frac{1}{\numG} \sum_{i=1}^{\numR}\frac{\ball{U}{R}{k}{G}}{\ball{U}{R}{k}{R}+1}
\end{equation}
\begin{equation}
{J}_{1}(\rg, \rr) = \mathbb{{E}}_{x\sim\rr}\left[\frac{f_{\rg}(x)}{f_{\rr}(x)}\right] = 1.
\end{equation}
The difference between Density and $\widehat{J}_{1}(G, R)$ is the added 1 in the denominator, a shared difference with that of $PC$ and $\widehat{J'_{1}}(R, G)$. We highlight the fact that ${J}_{1}(\rg, \rr) = 1$, which implies that as $\numG$ increases, and with $k$ a growing function of $\numG$ \cite{noshad2017direct}, Density and $\widehat{J}_{1}(G, R)$ are both asymptotically 1. Because of differing convergence rates, we note that Density can still be a meaningful measure when comparing output with the same number of drawn samples. However, as our Appendix experiments show, Density is an unreliable measure when working with an increasingly large number of samples.

\textbf{Intuition}.
Here, we describe the intuition behind why the Density metric is asymptotically 1 under appropriate conditions.
For clarity, we present definitions of $J_{1}(\rg, \rr)$ and its estimator $\widehat{J}_{1}(G, R)$ that are simply rearrangements of their formulae in Section 3:
\begin{equation}
J_{1}(\rg, \rr) = \int f_{\rr}(x)\frac{f_{\rg}(x)}{f_{\rr}(x)} d x = 1
\end{equation}
{\small
\begin{equation}
\widehat{J}_{1}(G, R)=\frac{1}{\numR} \sum_{i=1}^{\numR}\frac{\numR\cdot\ball{U}{R}{k}{G}}{\numG\cdot(\ball{U}{R}{k}{R}+1)}.
\end{equation}
}
Monte Carlo integration $\frac{1}{N_X} \sum_{i=1}^{\numX} f_{\gd}(X_i)$ serves as an unbiased estimator for $\int f_{\gd}(x) f_{\rd}(x) \, dx$. Applying this principle, $\frac{1}{N_R} \sum_{i=1}^{N_R} \frac{f_{\rg}(R_{i})}{f_{\rr}(R_{i})}$ is an unbiased estimator of $J_{1}(\rg, \rr)$ (Eqn. 20). Dissecting $\widehat{J}_{1}(G, R)$, the average $\frac{1}{\numR} \sum_{i=1}^{\numR}$ approximates $f_{\rr}(x)$, while the scaled ratio of $k$NN points approximates the probability density ratio $\frac{f_{\mathcal{G}}(R_{i})}{f_{\mathcal{R}}(R_{i})}$. Combining these two ideas, $\widehat{J}_{1}(G, R)$ can be constructed to estimate $J_{1}(\rg, \rr)$, which is equal to 1. We therefore affirm that $\widehat{J}_{1}(G, R)$ and density asymptotically approach 1. 

\textbf{Connecting Coverage}. We observe that Coverage is a special case of Recall Coverage (see the empirical $(k, k')$-$\mathit{PRC}$ definition earlier and Eqn.~\ref{eqn:RC}), where $C=k$ and $k'=1$. Hence, leveraging $k=\#(R, B_{k,R}(R_i))$, we can define Coverage as: 
\begin{equation}
\text { Coverage }=\frac{1}{\numR} \sum_{i=1}^{\numR} \textbf{1}\big[\frac{\#(G, B_{k,R}(R_i))}{\#(R, B_{k,R}(R_i))}\ge \frac{1}{k}\big].
\end{equation}
Thus, Coverage is also an estimator for $J'_{1}(\rg, \rr)$. 

\subsection{Implication} 
As demonstrated from our mathematical analysis, these metrics approximate the general form of a statistical divergence $D(\rd || \gd) = \mathbb{{E}}_{x\sim\rd}\left[g\left(\frac{f_{\gd}(x)}{f_{\rd}(x)}\right)\right]$, where $PC$, $RC$, and Coverage use $g(z) = \textbf{1}[z \ge \frac{\eta}{C}]$, Density uses $g(z) = z$, and KL-divergence is $g(z) = -log(z)$. 
We observe that in the context of precision, the form $D(\rg || \rr)$ is ineffective due to its dependence on relative densities. For instance, $J'_{1}(\rg, \rr)$ (related to $PC$) includes a $f_\rg(x)$ term in the denominator, which unfairly punishes for when the generated region is dense --- an aspect which should be of neutral relevance.  
In the context of recall, the form $D(\rr || \rg)$ is more reliable. Regions with high probability in the real distribution should correspondingly require dense coverage by generated samples; conversely, dense generated regions should not necessitate denser real regions to be considered precise. 

We substantiate our observations, noting $D_{KL}(\rg\|\rr) = CE(\rg, \rr)-H(\rg)$, by comparing the closed-form expression for all three terms when increasing the variance of $\rg$. We analyze two multivariate Gaussian distributions, $\rr \sim \mathcal{N}(\boldsymbol{0}, \mathbf{I})$ and $\rg \sim \mathcal{N}(\boldsymbol{0}, \sigma^2\mathbf{I})$ in $\mathbb{R}^{10}$, where $\boldsymbol{0}$ represents the zero mean vector and $\mathbf{I}$ denotes the identity matrix.
\begin{table}[h]
\centering
\label{tab:kl_divergence}
\begin{tabular}{|c!{\vrule width 1pt}c|c|c|}
\hline
\textbf{$\sigma^2$} & \textbf{$D_{KL}(\rg\|\rr)$} & \textbf{$CE(\rg, \rr)$} & \textbf{$H(\rg)$} \\ 
\thickhline
0.25 & 3.18 & 10.44 & 7.26 \\ \hline
1 & 0.00 & 14.19 & 14.19 \\ \hline
2.5 & 2.92 & 21.69 & 18.77 \\ \hline
\end{tabular}
\caption{KL-divergence $D_{KL}(\rg\|\rr)$, cross-entropy $CE(\rg, \rr)$, and entropy $H(\rg)$ for varying values of $\sigma^2$, the variance of the generated distribution of $\rg$.}
\end{table}

Precision measures should observe decreased fidelity as $\rg$ increases its variance relative to that of $\rr$. Notably, the $D_{KL}$ values at $\sigma^2 = 0.25$ and $2.5$ are approximately symmetric around the $\sigma^2 = 1$ case, illustrating how both underestimation and overestimation of variance relative to the true model contribute comparably to the divergence. We highlight that the basis of our metric defined in Section 5, $CE(\rg, \rr)$, trends in the right direction. These insights motivate the construction of our asymmetric cross-entropy measures.
\section{Information-theoretic Precision and Recall Metric}
\label{sec:our_metrics}
Here, we introduce the three components of our comprehensive metric: Precision Cross-Entropy ($\pce$), Recall Cross-Entropy ($\rce$), and Recall Entropy ($\re$). 
\subsection{Desiderata for Generative Model Metrics}
\label{sec:metrics_char}
To clarify the optimal behavior of generative model metrics,
we first present the following desiderata for precision and recall measures.

\textit{Precision} should (P1) reflect high fidelity for generated 
samples that are likely within the real distribution, (P2) indicate 
low fidelity for those that are improbable, highlighting cases of 
mode invention~\cite{borji2018prosconsganevaluation}. For certain domains such as image generation, where realism is effectively judged by human evaluation, precision should (P3) align with human evaluation \cite{stein2024exposing}. 

\textit{Recall} should (R1) reward the case where high-probability regions of the real distribution are sufficiently covered by generated samples, addressing mode dropping, and where (R2) the generated samples capture the full variation within each mode, limiting mode shrinkage. 

\subsection{Theoretical Metric Definitions} 
We exploit the asymmetric nature of (cross-)entropy, as defined in 
Section 3.1, to construct an information-theoretic
interpretation of precision and recall. Below, we define our metric
and justify its use in accordance with the labeled desiderata in Section~\ref{sec:metrics_char}. 

The three components of our metric are defined below. Note that for all components, the entropy of the real data distribution, $H(f_{\rr})$, serves as a constant baseline, so that for identical distributions, $\pce = \rce = \re = 0$. Because $H(f_{\rr})$ is independent of the generated distribution, the quality of the metric scores does not depend on its value.
\newline
\newline
\textbf{Precision Cross-Entropy ($\pce$):} 
\begin{align}
\pce &= CE(\rg, \rr) - H(\rr)\nonumber\\
&= \mathbb{E}_{x\sim\rg} \log \frac{1}{f_{\rr}(x)}-\mathbb{E}_{x\sim\rr} \log \frac{1}{f_{\rr}(x)}\,. 
\label{eqn:pce}
\end{align}
$\pce$ is minimized 
when generated points are in high-probability regions of the real
manifold~(P1). Lower values of $\pce$ correspond to higher levels of fidelity and less mode invention~(P2). 

\noindent
\textbf{Recall Cross-Entropy ($\rce$):}  
\begin{equation}
\label{eqn:rce}    
\rce = CE(\rr, \rg) - H(\rr) = \mathbb{E}_{x\sim\rr} \log \frac{f_\rr(x)}{f_{\rg}(x)}\,.
\end{equation}
Minimizing $\rce$ requires that the regions where the real probability is large also exhibit large values of the generated probability~(R1). Low values of $\rce$ indicates high inter-class diversity and therefore limited mode dropping. Note that $\rce$ is equivalent to $D_{KL}(\rr||\rg)$.

\noindent \textbf{Recall Entropy ($\re$):} 
\begin{align}
\re &= H(\rg) -  H(\rr)\nonumber\\
&= \mathbb{E}_{x\sim\rg} \log \frac{1}{f_{\rg}(x)} - \mathbb{E}_{x\sim\rr} \log \frac{1}{f_\rr(x)}\,. \label{eqn:re}
\end{align}
The more similar generated points are to each other (i.e. \textit{lower} uniqueness), the \textit{lower} $\re$ is. Low $\re$ coupled with low $\pce$ indicates mode shrinkage, as points clustered at the average of a mode are highly precise, but not unique (R2). $\re$ can be seen as the intra-class diversity complement to $\rce$. 
$\re$ and $\pce$ are normalized versions of the two  components of $D_{KL}(\rg\|\rr) = CE(\rg, \rr)-H(\rg)$.



\subsection{Empirical Definitions}
\label{sec:inftheory_empirical}
For calculations on sample sets, we refer back to the 
\textit{k}NN-based estimators defined in Section~\ref{sec:knn} (see Eqn. 3 and 4)
and define the empirical version of our metric as: 
\begin{align}
\pce_k &= \breve{CE}_k(G, R)-\breve{H}_k(R)\\
\rce_k &= \breve{CE}_k(R, G)-\breve{H}_k(R) \\
\re_k &= \breve{H}_k(G)-\breve{H}_k(R).
\end{align} 
From a high-level perspective, $\pce_k$ calculates the average log-normalized distance to the \textit{k}th-nearest neighbor in the real set for all generated samples. 
Similarly, $\rce_k$ is minimized when the average normalized distance 
of real points to their generated neighbors is small. Conversely, 
$\re_k$ is sensitive to when generated points are clustered together. In this empirical context, $\breve{H}_k(R)$ can also be seen as a
normalization factor dependent on the dimension and choice of $k$ \cite{Park_2023}. 
These scores leverage the same principle as other \knn metrics and $FD$;
distances in a feature space convey meaningful information, with 
shorter distances signifying greater similarity between points.  

Unlike non-\textit{k}NN-based metrics, such as $FD$, our scores can be dissected for mode- and sample-level analysis, as the estimators are additive in the contributions of different data points. Hence, we can see which samples contribute the most to the overall
score and what modes the generative model is struggling to represent. We note that in cases where $\pce_k$ is significantly negative for a sample, 
memorization---when the model generates points (near-)identical to a 
training sample---may be present. Examples are provided in the Appendix. 

Because of its favorable sample-level nature, our method can be easily related to other works. For instance, generated samples can be sorted by contribution to $\pce_k$, and the highest scoring can be discarded; this is a form of model-auditing \cite{alaa22synthetic}, where samples are judged for quality and the most unrealistic samples are rejected. We can also extend our method to Precision-Recall curves \cite{djolonga2020precision}, where we can visualize a trade-off between $\pce_k$ and $\re_k$ when auditing levels are increased. When the less realistic samples, which likely reside in the fringes of modes, are removed, intra-class diversity decreases while fidelity increases.

\section{Experiments}
\label{sec:experiments}

Here we analyze the empirical behavior of Density \& Coverage ($D$ \& $C$), Precision Recall Cover ($PC_{k, k'}$ \& $RC_{k, k'}$), \frechet distance (\fd), and our metric ($\pce_k$, $\rce_k$, \& $\re_k$), specifically focusing on how well they align with the desiderata listed in Section \ref{sec:metrics_char}. We set $k$ to 5, $k'$ = 15, and omit the subscripts from the metric abbreviations for brevity hereafter. We follow the recommendation of \citet{stein2024exposing} to embed the images with the DINOv2-ViT-L/14 encoder \cite{oquab2024dinov2learningrobustvisual}, which they claim provides a richer representation space than the commonly used Inception network, which may unfairly punish diffusion models. This motivates our generalized abbreviation $FD$.

\textbf{Dataset Descriptions.} We use both ImageNet \cite{deng2009imagenet} and CIFAR-10 \cite{krizhevsky2009learning} image datasets for our analysis. The sampled training set for ImageNet contains 1000 classes with 100 images each, while CIFAR-10 has 10 classes with 4500 images each --- more information about dataset usage is available in the Appendix.
\begin{figure}[t]
\includegraphics{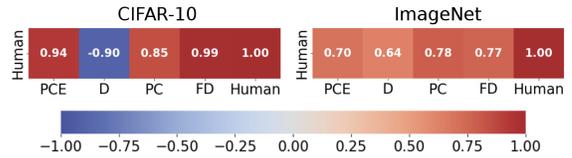}
\centering
\caption{Pearson correlation heatmap between the precision measures and human error rate, displayed separately for the CIFAR-10 and ImageNet datasets. The direction of the scores where higher values indicate poorer precision was flipped so that all metric scores rank in the same direction. } 
\end{figure}

\textbf{Correlation with Human Rankings}. Although there is no clear ``ground truth'' for overall grading of generative output, human assessment can be seen as the gold standard for the realism component of metrics. 
\citet{stein2024exposing}
recently evaluated human perception for a multitude of models, using a similar method to the HYPE metric mentioned in Section 2, creating a meaningful baseline for realism scores. Using their recommended DINO-v2 space, they find that only $FD$ correlates well with human assessment, with their experiments including density and improved precision \cite{kynkaanniemi19precision}. We extend their experiments on the CIFAR10 and the more complex ImageNet datasets, adding $PC$ and our precision measure $\pce$ to the analysis. 
As seen in Figure 2, we find that $\pce$, $PC$, and $FD$ all have a reasonably high and similar ($\geq 0.7$, all within a range of $0.08$ units) positive correlation with human error rate for both datasets. Although Density ($D$) shows a positive correlation on the ImageNet dataset, it shows a heavy negative score on CIFAR-10. This inconsistent relationship with human error rate highlights a significant weakness of the Density measure when measuring realism on practical datasets.

\textbf{Mode Shrinkage Experiment}. The classifier-free guidance (CFG) parameter \cite{ho2022classifier} allows control over trading off intra-class diversity for fidelity. Increasing CFG leads to datasets with worse mode shrinkage but higher precision, similar in effect to Figure 1(c). Adapting the diversity experiment from \citet{stein2024exposing}, we calculate the metrics for image sets generated at five levels of the CFG parameter, steadily decreasing the intra-class diversity of the generated output, as shown in Figure 3. The DiT-XL-2 model, trained on ImageNet, was used for generation. With increasing CFG, precision metrics should show increased fidelity, while recall---specifically $\re$ for our metric--- should show decreased diversity. However, we find that $PC$ decreases for higher CFG values, while $D$ and $\pce$ perform as expected. The recall metrics ($C$, $RC$, $\rce$) see only a slight change - only our intra-class diversity metric ($\re$) shows significant sensitivity to mode shrinkage. 
\begin{figure}[t]
\centering
\includegraphics{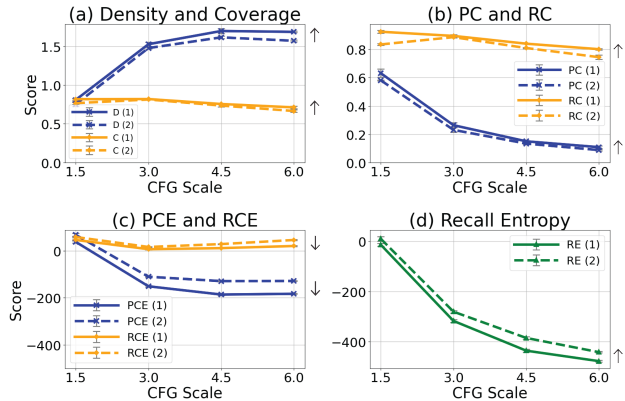}
\caption{Behavior of metrics at different CFG values for the generation of samples using the DiT-XL-2 model trained on ImageNet. Two sets of 15 random classes were used (as marked by (1) and (2)), with six runs each performed and averaged. Error bars (±1 SD) are plotted but are visibly negligible. Arrows are put aside each measure's lines - upright arrows mean that higher values indicate better fidelity or diversity (opposite for the downward arrows). See sample images in the Appendix and Figure 1(c). } 
\end{figure}

\textbf{Mode Dropping Experiment}. To complement our intra-class diversity test, we conduct an inter-class diversity experiment. We start with the original 1000 classes from ImageNet, with 100 images generated per class by the ADMG-ADMU \cite{dhariwal2021diffusion} model. 
As seen in Figure 4, 100 classes were dropped at a time until 100 classes remained, for a total of 10 intervals. Ideally, mode dropping metrics ($C$, $RC$, $\rce$) should show significant change, while precision and mode shrinkage measures ($D$, $PC$, $\pce$, $\re$) should remain relatively constant. We find that to be the case. Figure 3(d) shows that the one-dimensional $FD$ cannot distinguish diversity loss from precision, highlighting the benefit of multi-dimensional metrics.

\begin{figure}[h]
\includegraphics{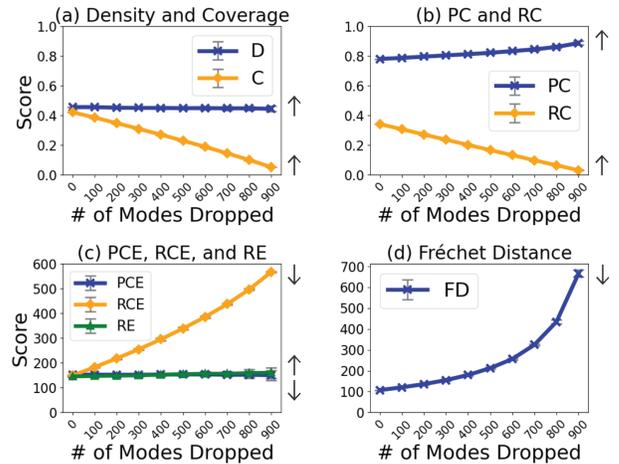}
\centering
\caption{100 random classes were dropped at a time from the ImageNet ADMG-ADMU generated dataset. The corresponding metric scores are plotted. The scores of 10 runs were averaged and (±1 SD)  error bars are plotted, though visually minimal. Similar to Figure 3, arrows are beside each measure's lines, with the orientation of the arrow corresponding to the direction in which better fidelity/diversity is defined.} 
\end{figure}

\textbf{Summary}. We find through our experiments that:
\begin{itemize}
  \item Unlike metrics under the precision and recall framework, \frechet distance ($FD$) cannot differentiate between failure modes, as per its one-dimensional nature.
  \item Our inter-class diversity measure, Recall Cross-Entropy ($\rce$), Coverage ($C$) and Recall Coverage ($RC$), are all highly sensitive to mode dropping, while their precision counterparts remain relatively constant.
  \item Out of the recall measures, only our intra-class diversity measure, Recall Entropy ($\re$), is significantly sensitive to mode shrinkage.
  \item Only our Precision Cross-Entropy ($\pce$) both correctly correlates with human perception and increases with mode shrinkage. Density ($D$) fails to correlate in the correct direction for CIFAR-10, while Precision Coverage ($PC$) decreases as generated samples become more clustered.
\end{itemize}

\section{Concluding Remarks}
\textbf{Limitations}. Quality semantic embedding is needed for our metric to produce meaningful results. However, we emphasize that significant efforts across data modalities have been directed towards improving encoder capabilities and developing automatic, embedding-based evaluation metrics \cite{zhang2021automaticevaluationmoderationopendomain, huang2020tabtransformertabulardatamodeling}. Specifically, the wide-spread application of Fréchet distance, which requires an embedded space, to text, videos etc. \cite{xiang2021assessingdialoguesystemsdistribution,unterthiner2019accurategenerativemodelsvideo}, supports our metric’s application across data types. Another limitation and future research direction is expanding to data with more complex structures, such as graphs and trees, or to specialized generative tasks like surrogates of scientific simulations, where defining an appropriate probability distribution and a meaningful distance measure is nontrivial.

\textbf{Conclusion}. In this work, we show that some of the prior approaches to precision and recall metrics for generative
models can be reduced to nonparametric measures of information-theoretic divergence. This unifying
perspective also allows us to identify a significant gap in existing recall metrics: their limited 
sensitivity to intra-class diversity loss (mode shrinkage).
We then introduce a novel three-dimensional metric for evaluating generative models, comprised of 
Precision Cross-Entropy, Recall Cross-Entropy, and Recall Entropy. These components address three 
critical failure modes of generative models: mode invention, mode dropping, and the often-overlooked 
mode shrinkage.
Both empirical evidence, such as correlation with human judgment, and theoretical insights are used to 
demonstrate the advantages of our proposed metrics over other approaches.

\section*{Acknowledgements}
This work was supported by the USDA-NIFA/NSF AI Research Institutes program, under award No. 2021–67021–35344 and
by NSF Expeditions in Computing CCF-1918656.

\bibliography{aaai25}

\clearpage

\appendix

\section{Resources and Reproducibility}
The computing needs of this project were modest, so we used either Google Colab Pro+ or our host instituition's high-performance computing environment. The latter involved CPUs with 40-96 cores and 375-750 GB memory per Intel or AMD node. When needed, we used A100 GPUs on both enviroments. We use an environment with Python 3.11.4, Pytorch 2.2.0, Cuda 12.1, and scikit-learn 1.3.0. We provide a zip file with code for our experiments and as the full datasets are too large, we upload representative samples. Full datasets are publicly available through links on our GitHub, together with the code and documentation, made available under the Apache license. 
We base our experiments on the authoritative datasets, CIFAR-10 and ImageNet, as they have an excellent range of levels of resolution, number of classes, and number of samples per class. We used datasets from 13 models trained on CIFAR-10 and 9 models trained on ImageNet, publicly provided by \citet{stein2024exposing}, to perform the human judgement correlation experiments. Further information about models used and image pre-processing can be found in Appendix A of their work.  Using data from their human subject experiments, \citet{stein2024exposing} define a human error rate for each model to create a ranking baseline for precision measures. We also link their data in our GitHub: https://github.com/NSSAC/PrecisionRecallMetric.

\section{Bias Expression for Metric Estimates}

Here, we extend the bias derivation for the nearest neighbor ratios estimator~\cite{noshad2017direct} to be applicable to $PRC$ and Density \& Coverage, defined in Section 4. We define modifications of $\widehat{J}_{\alpha}(X, Y)$ below to derive the bias expression. There are three key differences between the metric formulae and $\widehat{J}_{\alpha}(X, Y)$: as in the original analysis, (i) the nearest neighbors are calculated with respect to $Z$ rather than the individual sets $X$ and $Y$, (ii) there is an additional 1 in the denominator, and (iii) there is no indicator function. 

For clarity, we define $k + k' = K$, where for a given ball around a point, $k$ of the neighbors are from $X$, $k'$ are from $Y$, and the total number of neighbors in the union set $Z$ is $K$. 
We define $\widehat{J}_{\alpha}^{*}(X, Y)$ as a modification of $\widehat{J}_{\alpha}(X, Y)$, where the nearest neighbors within the ball are calculated relative to points in $Y$ rather than those in $Z$:
\begin{equation}
\widehat{J}_{\alpha}^{*}(X, Y)=\frac{\eta^{\alpha}}{\numY} 
\sum_{i=1}^{\numY}\Big(\frac{\ball{Y}{Y}{k'}{X}}{\ball{Y}{Y}{k'}{Y}}\Big)^{\alpha}, \label{eqn:Jalpha*}\\
\end{equation}
The general idea behind the following steps is that if it can be proven that $K$ does not need to be the same for all samples, and there is a set of $K(i)=[K_{1}, K_{2}...K_{\numY}]$ such that $k'$ is constant across $i$, then a bias term can be derived for the modified estimator. 
\\To prove using a set of $K(i)$ works similarly to fixed $K$, we show that the bias proof works for sample-level density ratio estimates. 
We refer to Equation 14 from \cite{noshad2017direct}, which expands the the expectation function $\mathbb{E}[\widehat{J}_{\alpha}(X, Y)] $:
\begin{align}
\mathbb{E}[\widehat{J}_{\alpha}(X, Y)] & =\frac{\eta^{\alpha}}{\numY} \mathbb{E}\bigg[\sum_{i=1}^{\numY}\Big(\frac{\ball{Z}{Y}{K}{X}}{\ball{Z}{Y}{K}{Y}+1}\Big)^{\alpha}\bigg] \label{eq:eJalpha}\\
& =\eta^{\alpha} \mathbb{E}_{Y_{1} \sim \gd} \mathbb{E}\bigg[\Big(\frac{\ballone{Z}{Y}{K}{X}}{\ballone{Z}{Y}{K}{Y}+1}\Big)^{\alpha} \rvert\, Y_{1}\bigg].
\end{align}
In Eqn. \ref{eq:eJalpha}, $\sum_{i=1}^{\numY}$ can be moved to the outside because of the linearity of expectations ($\mathbb{E}[X+Y]=\mathbb{E}[X]+\mathbb{E}[Y]$). Therefore, (\ref{eq:eJalpha}) can be rewritten as 
\begin{equation}
\mathbb{E}[\widehat{J}_{\alpha}(X, Y)] =\frac{\eta^{\alpha}}{\numY}\sum_{i=1}^{\numY}\mathbb{E}\bigg[\Big(\frac{\ball{Z}{Y}{K_{i}}{X}}{\ball{Z}{Y}{K_{i}}{Y}+1}\Big)^{\alpha}\bigg].
\end{equation}
Applying the Law of Total Expectation ($\mathbb{E}_{X}[X] = \mathbb{E}_{Y}[\mathbb{E}_{X}[X|Y]]$), we write
\begin{equation}
\small
\mathbb{E}[\widehat{J}_{\alpha}(X, Y)] =\frac{\eta^{\alpha}}{\numY}\sum_{i=1}^{\numY}\mathbb{E}_{Y_{i} \sim \gd}[\mathbb{E}[(\frac{\ball{Z}{Y}{K_{i}}{X}}{\ball{Z}{Y}{K_{i}}{Y}+1})^{\alpha}|Y_{i}]].
\label{eq:eJalpha2}
\end{equation}
We introduce $\mathbb{E}[\widehat{J}_{\alpha, i}(X, Y)]$, which is the expectation of the sample-level density ratio estimate at point $X_i$. Condensing the interior expectation under the new notation, (\ref{eq:eJalpha2}) becomes
\begin{equation}
\mathbb{E}[\widehat{J}_{\alpha}(X, Y)] =\frac{\eta^{\alpha}}{\numY}\sum_{i=1}^{\numY}\mathbb{E}[\widehat{J}_{\alpha, i}(X, Y)].
\end{equation}
\\The authors define the bias for $\widehat{J}_{\alpha}(X, Y)$ as:
\begin{equation}
\mathbb{B}[\widehat{J}_{\alpha}(X, Y)]=O\Big(\big(\frac{K}{\numX}\big)^{\gamma / d}\Big)+O\big(\frac{1}{K}\big),
\end{equation} where $\gamma$ is parameter for the Hölder smoothness class, explicitly defined in Eqn. 3 of their work.

The $O((\frac{K}{\numX})^{\gamma / d})$ term originates from Lemma 3.1 of~\citet{noshad2017direct}, which bounds the local smoothness of the density functions. Since the lemma is defined locally, the bound can be used on a sample-level basis. Similarly, the $O(\frac{1}{K})$ term, which comes from the de-Poissonization of the estimator, can be separated for each $Y_i$ term, and therefore is also applicable on a sample-level basis. 

We derive the bias for $\widehat{J}_{\alpha}^{*}(X, Y)$:

$$
\mathbb{B}[\widehat{J}_{\alpha,i}(X, Y)]=O\Big(\big(\frac{K_{i}}{\numX}\big)^{\gamma / d}\Big)+O\big(\frac{1}{K_{i}}\big) 
$$
\begin{equation} \mathbb{B}[\widehat{J}_{\alpha}^{*}(X, Y)]=\frac{1}{\numY}\sum_{i=1}^{\numY}\bigg[O\Big(\big(\frac{K_{i}}{\numX}\big)^{\gamma / d}\Big)+O\big(\frac{1}{K_{i}}\big)\bigg]. 
\end{equation}

$\frac{1}{\numY}\sum_{i=1}^{\numY}O((\frac{K_{i}}{\numX})^{\gamma / d})$ can be bounded by $O((\frac{K_{max}}{\numX})^{\gamma / d})$. 
$\frac{1}{\numY}\sum_{i=1}^{\numY}[O((\frac{1}{K_i})$ can be bounded by \(O(\frac{1}{K_{min}})\). Therefore,
\begin{equation}
\mathbb{B}[\widehat{J}_{\alpha}^{*}(Y, X)]=O\Big(\big(\frac{K_{max}}{\numX}\big)^{\gamma / d}\Big)+O\big(\frac{1}{K_{min}}\big).
\label{eqn:bias_estimate}
\end{equation}
In the formula, $K_{min}$ is the minimum value for $K=k+k'$ over all samples. $K_{min}$ is bounded below by $k'$, as always $k\ge0$. Therefore, we can replace $K_{min}$ by $k'$ above. Next, we define $K_{max}$ as the maximum value for $K$ over the samples --- here we need an upper bound to use in Eq. 35. For extreme distributions, this can be problematic, as there could be a large number of points in $X$ and $Y$ clustered together. This would lead to large values of $K_{max}$, and therefore, a poor bias estimate. However, \textit{k}NN-based estimates are marginally useful whenever the distribution of points has extreme peaked behavior.

We then define a second modification to $\widehat{J}_{\alpha}(X, Y)$, adding 1 to the denominator:
\begin{equation}
\widehat{J}_{\alpha}^{**}(X, Y)=\frac{\eta^{\alpha}}{\numY} 
\sum_{i=1}^{\numY}\Big(\frac{\ball{Y}{Y}{k'}{X}}{\ball{Y}{Y}{k'}{Y}+1}\Big)^{\alpha}.
\end{equation}
For a $k'$NN ball defined around point $Y_i$, we can define a $(K_i-1)$NN ball, where ($K_i = k'_i + k_i$), that has the same radius. Therefore, asymptotic bias estimates are not by affected by this change. 

Finally, to connect $PRC$ and coverage, we need to add the indicator function. Since the indicator function is being used to check whether the argument is greater than a constant, we can view it as the limit of a sigmoid with a parameter $\theta$, which is asymptotically infinite, given by $\sigma(x; \theta) = \frac{1}{1 + e^{-\theta (x - \frac{\eta}{C})}}$ (see the discussion after Equation~\ref{eqn:J1primetheoretical}). The third modification can thus be defined as: 
\begin{equation}
\widehat{J}_{\alpha}^{***}(X, Y)=\frac{\eta^{\alpha}}{\numY} 
\sum_{i=1}^{\numY}\sigma\bigg(\Big(\frac{\ball{Y}{Y}{k'}{X}}{\ball{Y}{Y}{k'}{Y}+1}\Big)^{\alpha}; \frac{\eta}{C}, \theta\bigg).
\end{equation}
Applying Lemma 3.2 from~\citet{noshad2017direct}, the bias argument remains unchanged for finite $\theta$. The final expression for the bias is given by Eqn.~\ref{eqn:bias_estimate} for Density; $PRC$ and Coverage follow the same general form. 

\section{Additional Experiments}

\textbf{Increasing Sample Size}. We first perform a test of multivariate Gaussians to observe the behavior of Density as the number of samples increases. Our analysis (in Section 4) describes how density, under appropriate conditions, should asymptotically go to 1. We use $PC$ (Precision Coverage) as a baseline metric, as the two measures have similar scales. We define two Gaussian distributions: $\mathcal{R}$ with mean $\boldsymbol{\mu}_r = (0, 0)$ and $\mathcal{G}$ with mean $\boldsymbol{\mu}_g = (2, 2)$. Both distributions are generated with the same covariance matrix, $\Sigma = \begin{bmatrix} 1 & 0.1 \\ 0.1 & 1 \end{bmatrix}$. We visualize the two distributions used in Figure 5; evidently, there is substantial difference between them. 
However, as seen in Figure 6, Density starts at approximately $0.69$ and increases towards $y=1.0$; the bias term in Eqn. (35) implies bias decreases with increasing sample size. On the other hand, $PC$ is essentially constant, remaining within $0.015$ units from the beginning value. 

\begin{figure}[H]
\centering
\includegraphics[scale=0.2]{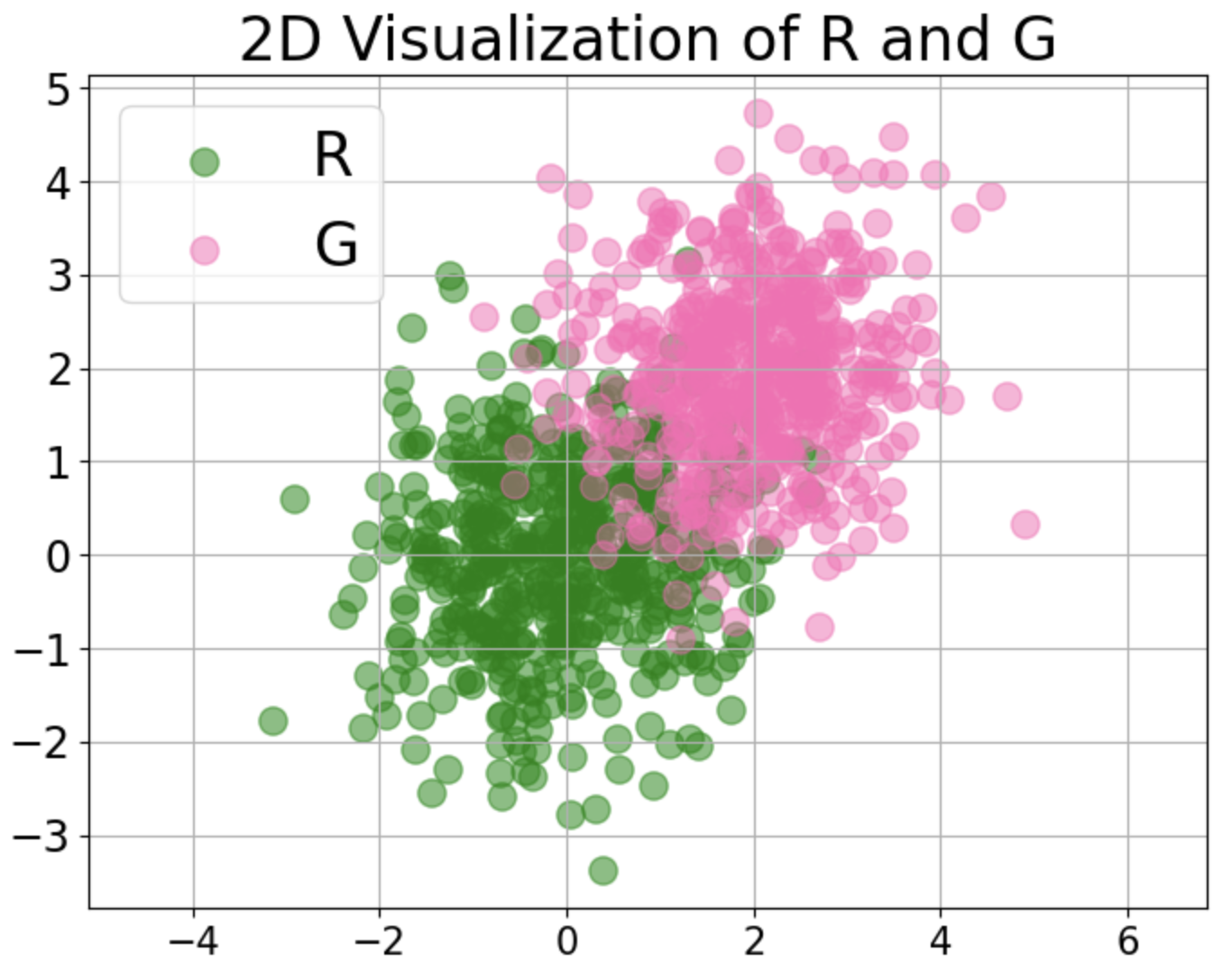}
\caption{Two-dimensional visualization of the sample set of distributions used for calculations in Figure 6.  } 
\end{figure}
\begin{figure}[H]
\centering
\includegraphics[scale=0.38]{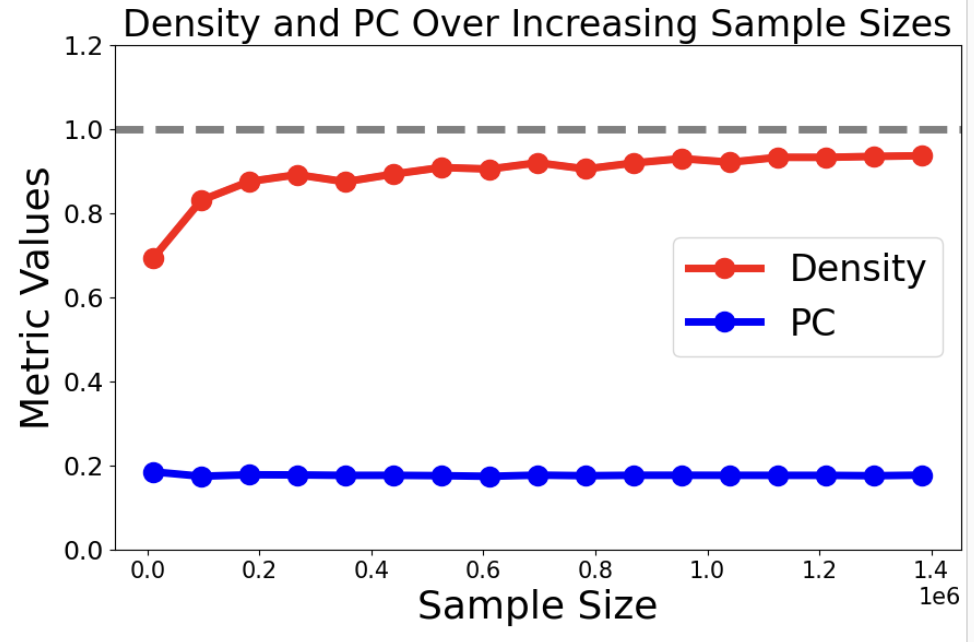}
\caption{Behavior of Density and $PC$ when sample size is substantially increased, from $1 \times 10^4$ to $1.5 \times 10^6$. The horizontal dashed line at $y=1.0$ signifies what our analysis claims density converges too. } 
\end{figure}

\textbf{Memorization}. As noted in Section 5.3, our measures can be dissected for sample-level analysis. Specifically, we mentioned that extremely low values of $PCE$ (in the negatives) can correspond to another failure mode, namely memorization. Negative $PCE$ values signify that the sample is closer to a real point than the average real point to its neighbor, which is an identifier for memorization. Following \citet{alaa22synthetic}'s idea, samples with extreme values can be audited to create a more authentic generated dataset. Certain models trained on the CIFAR-10 dataset are known to have memorization issues~\cite{stein2024exposing}, and while a small portion of samples are considered ``memorized,'' identifying this pathological case is important. Figure 7 illustrates how our metric can be utilized for this purpose.

\begin{figure}[H]
\centering
\includegraphics[scale=0.28]{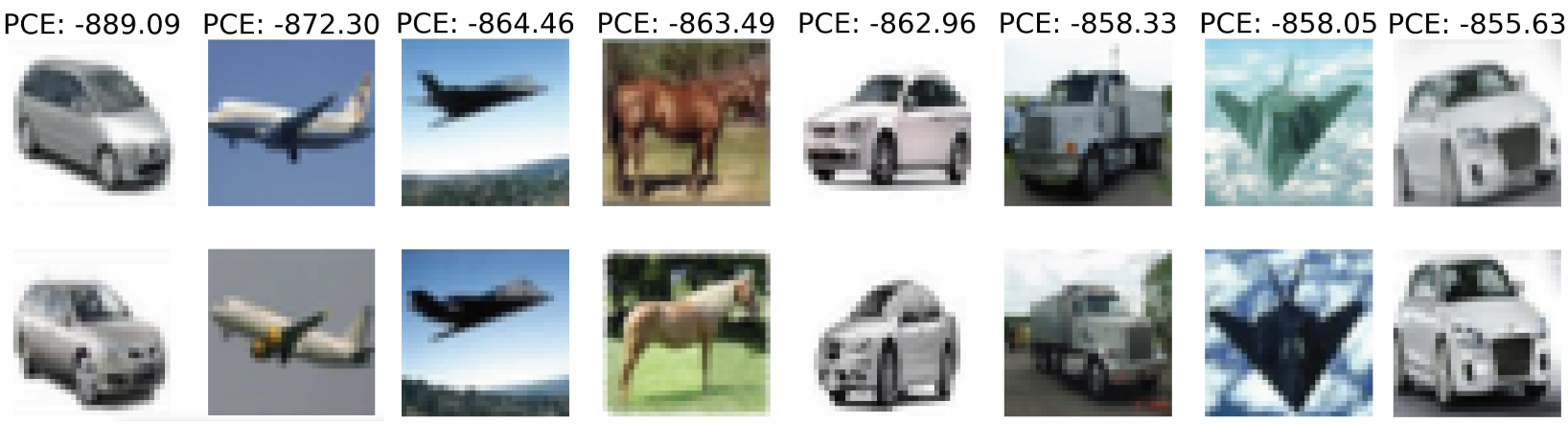}
\caption{ A sample of the ``most memorized'' images from a CIFAR-10 generated dataset. Each generated sample is annotated with its sample-level $PCE$ score above it. Below each is its nearest real neighbor. } 
\end{figure}

\end{document}